\documentclass[11pt]{article}
\usepackage{acl2015}
\usepackage{times}
\usepackage{url}
\usepackage{latexsym}
\usepackage{booktabs}
\usepackage{graphicx}

\title{Automated Extraction of Number of Subjects in Randomised Controlled Trials}

\author{Abeed Sarker \\
  Department of Biomedical Informatics \\
  Arizona State University \\
  Mayo Clinic, Scottsdale 85259, AZ \\
  {\tt abeed.sarker@asu.edu} \\ 
  }

\date{}

\begin{document}
\maketitle
\begin{abstract}
 We present a simple approach for automatically extracting the number of subjects involved in randomised controlled trials (RCT). Our approach first applies a set of rule-based techniques to extract candidate study sizes from the abstracts of the articles. Supervised classification is then performed over the candidates with support vector machines, using a small set of lexical, structural, and contextual features. With only a small annotated training set of 201 RCTs, we obtained an accuracy of 88\%. We believe that this system will aid complex medical text processing tasks such as summarisation and question answering.  
 
\end{abstract}

\section{Background}
The vast and growing volume of published medical literature has necessitated the need for automatic text processing systems. Natural language processing (NLP) research within the field of evidence-based medicine has focused on extracting key information from medical texts and applying them for automating complex, time-consuming tasks such as summarisation, question answering, and evidence appraisal \cite{demner07,mishra14review,sarker15grading}. However, primarily due to the complex nature of medical text, with its domain exclusive terminologies and ontologies, extraction of task-specific information is not trivial \cite{summerscales14thesis}. Problems of content selection/information extraction are perhaps further exacerbated by the varying structures of published medical texts \cite{altashared12}.

\begin{figure}[!h]
\centering
\fbox{
\scalebox{0.25}
{\includegraphics{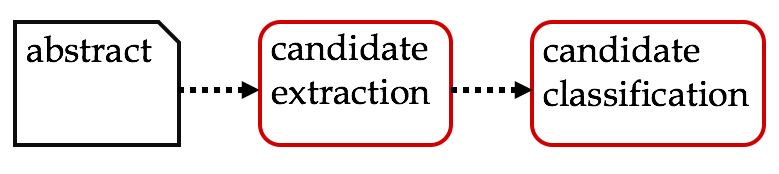}}}
\caption{Pipeline for the extraction of sample sizes from abstracts. }
\label{pipeline}
\end{figure}

In this paper, we present an approach for automatically extracting the sample sizes of randomised controlled trials. Our approach executes in two steps. In the first step, integers potentially representing study sizes are identified and extracted. The candidates are then classified using supervised support vector machine (SVM) classifiers trained on annotated size candidates from a set of 201 abstracts. Figure \ref{pipeline} shows the two steps involved in the process--- the first being a rule-based step, and the second being a supervised classification step. Our approach obtains an accuracy of 88\% on a blind evaluation set of 50 abstracts. Our experiments show promising results utilising a small annotated set, and using a mixture of lexical, structural and contextual features. This approach can be easily extended to other document types, such as other clinical trials and cohort studies. We also believe that the system will act as a platform for other information extraction tasks and a useful module in other domain-specific text processing tasks.

\section{Related Work}
There has been steady research on automatic information extraction tasks from medical publications, particularly for subsequent use in tasks such as question answering (\emph{e.g.}, Lin and Demner-Fushman (2007)\nocite{demner07}). Some of the specific information that have been of interest to information extraction problems are age values (values describing sample age characteristics), medical conditions, population groups (\emph{e.g.}, males, adolescents, elderly females), group sizes, outcomes, and so on \cite{summerscales14thesis}. Rule-based approaches have been heavily explored for these and similar tasks--- the majority of approaches attempting to exploit lexical regularities, syntactic structures, and/or other surface-level lexical properties (\emph{e.g.}, Summerscales \emph{et al.} (2011)\nocite{summerscales11}, and Bruijn \emph{et al.} (2008)\nocite{bruijn08}). The primary problem faced by rule-based systems is extensibility, as the number of rules can become very large, and increasingly difficult to maintain \cite{chiticariu13}. Supervised learning approaches have been shown to be more effective in the past \cite{hansen08,kiritchenko10}, however their portability and availability for use in more complex systems have not been assessed. Relative to past work, the problem we attempt to address is not novel. However, from an application perspective, the intent of our work has two primary distinctions: (i) we explore the effects of data-centric and deep linguistic features in a hybrid framework, (ii) our implementation is particularly aimed at portability with the target \emph{end-users} being complex NLP systems, rather than humans.

\section{Methods}
\subsection{Data}
We collected a set RCTs from two sources: approximately 200 were randomly selected from all RCTs in the corpus proposed by Moll\'a and Santiago-martinez (2011), and an additional  \nocite{molla11} 200 were collected from PubMed.\footnote{\url{http://www.ncbi.nlm.nih.gov/pubmed}. [Accessed: 8-10-2015.]} RCTs were identified using the rule-based approach proposed by Sarker and Molla (2010) \nocite{sarker10}. RCTs not involving human subjects were removed, and a total of 251 RCTs were annotated. We found the manual identification of the sizes of studies relatively straightforward, as also suggested by Summerscales (2014).
\subsection{Candidate extraction and classification}

All integers from each abstract are extracted and considered as candidates. Integers expressed in words are first converted to their numeric forms using regular expressions and a lookup table. Integers lower than 10 are discarded. Note that while there might be RCTs with lower than 10 samples, we found that adding this criterion significantly lowers the number of potential false positives.\footnote{There were no studies of size smaller than 10 in our sample.} 

Following the extraction of candidates, a set of features are extracted associated with each candidate. The features can be broadly categorised into three groups, which are as follows:
\subsubsection{Contextual features}
\begin{itemize}
                       
\item \textbf{ Context terms.} For a candidate token in position \emph{n} in a sentence, we include the terms from positions \emph{n-3} to \emph{n+3} as features. We also incorporate the full context string (from \emph{n-3} to \emph{n+3}) as a separate single feature. All tokens are preprocessed by lowercasing and stemming. 

\item \textbf{Context clusters.} For each term in the context window, we introduce as feature a cluster number representing this token. The intent of using these cluster numbers is to allow a more generic representation of the context terms, which, we believe, may be particularly useful when we have a small training set. To generate the word clusters, we employ the popular word2vec tool\footnote{\url{https://code.google.com/p/word2vec/}. [Accessed: 8-17-2015].} \cite{mikolov13b}. Distributed representations of the words are first learned from a large set of approximately over 800,000 RCTs obtained from PubMed using simply the keywords: \emph{randomized controlled trial}. The only preprocessing performed is the lowercasing of the words. Following the generation of the distributed word vectors, K-means clustering \cite{macqueen67} is used to allocate the word vectors into 500 clusters.\footnote{We did not perform further experimentation to explore the possibility of better cluster allocations.} For each context term, its cluster number is used as a feature. 

\item \textbf{ Other context features.} During annotation, we observed that candidates close to the mentions of population groups are more likely to be study size mentions. At the same time, a number of candidates are descriptions of temporal concepts (\emph{e.g.}, `37 years old', `after 65 days'). We used features to capture these kinds of information. First, we prepared a set of terms representing population group mentions. We performed this by extending the list of terms suggested by Xu \emph{et al.} (2007)\nocite{xu07}. For each candidate, we used a binary feature to indicate if there is a population term present in the context window, and a numeric feature to represent the distance from the population term. As for temporal mentions, we used a binary feature to indicate if a time-indicating term is next to the candidate.
\end{itemize}

\subsubsection{Lexical features}
\begin{itemize}
\item \textbf{Sentence n-grams.} We generate 1-,2-,3-grams from the candidate sentence and use them as features. The text is preprocessed in the previously mentioned manner before generating the n-grams.

\item \textbf{Indication of year.} A large number of the candidates generated by our first processing step are year mentions (\emph{e.g.}, 1995, 2014 and so on). So, we tag all candidates between the numerical values of 1950 to 2020 with a binary feature indicating that the mention can possibly be a year.
\end{itemize}

\subsection{Structural features}
\begin{itemize}
\item \textbf{Sentence category.} For structured abstracts obtained from PubMed, sentences are often categorised into broad categories such as \emph{Methods}, \emph{Results}, \emph{Conclusions}, and so on. When available, we use these categories as features for the candidates.

\item \textbf{Sentence label.} Similar to categories, but the labels often represent more fine-grained information (\emph{e.g.}, \emph{Patients and methods}). We use these as candidate features. In addition, we identified some labels that are very likely to be contextually associated with size mentions of studies. We added a binary feature to indicate if the sentence associated with a candidate belongs to any of these highly likely labels.
 
\item \textbf{Candidate position.}
We use the absolute and relative candidate positions in the sentence as numeric features.
\item \textbf{Sentence position}
We use the absolute and relative sentence positions for a candidate as numeric features.

\end{itemize}
\subsection{Classification experimental setup}
Using these features, we use SVMs to perform classification. We use the 201 abstracts in the training set to automatically extract candidates, and then perform training using these automatically extracted candidates. Note that this form of training makes the future preparation of larger data sets simpler (\emph{i.e.}, the annotator is just required to tag the correct candidate extracted as a positive label). In total, 1,530 candidates are generated from our full training set. SVMs are set up with an \emph{RBF} kernel and the cost and gamma parameters are optimised via grid search, using 100\% of the train set for learning and evaluation. When performing parameter optimisation, the accuracies of individual candidate classifications are not considered. Instead, we define the classification problem as that of selecting the right candidates with the constraint that only one candidate per abstract may be chosen as the size. As such, our classifier generates probability estimates for each candidate, and chooses the candidate with the highest probability within an abstract as the size for that abstract. The best cost/gamma configuration is employed on the blind test set. 

Following the classification, we performed feature analysis on the three broad feature categories. The results and a brief discussion are provided in the next section.

\section{Results and discussions}
Table \ref{tab1} presents the performance of our system on the small test set, and the contributions of the individual types of feature sets. The best accuracy obtained on the test set is 88\%. The table also presents the confidence intervals, which are rather wide due to the small size of the test set. The best accuracy score obtained on the training set using 10-fold cross validation is 94\%. The table also presents the reported performances of several systems. Rather surprisingly, early statistical approaches presented very good results using relatively small data sets. However, newer approaches have reported performances much lower than ours. Because all these systems were evaluated on distinct data sets, and because none of these systems are publicly available, it is not possible to analyze the reasons behind the wide range of performances reported.

We also performed single feature set experiments and leave-out feature set experiments. For these, we considered each of the broad categories of features. Table \ref{tab1} shows that for the single feature scores, the contextual features perform the best, and the lexical features perform poorly when structural and contextual information are removed. The leave-out experiment results further verify this finding, and also reveal that lexical features help to boost performance.

\begin{table}
\center
\begin{tabular}{@{}lcc@{}} \toprule 
\textbf{Features} & \textbf{Accuracy (\%)} & \textbf{95\% CI}\\ \midrule \midrule
All& 88 & 76 -- 95 \\ 
Contextual& 80 & 66 -- 90 \\ 
Structural & 76 & 62 -- 87 \\ 
Lexical & 12 & 4.5 -- 24\\
- Contextual& 82 & 69 -- 91 \\ 
- Structural & 80 & 66 -- 90 \\ 
- Lexical & 84 & 71 -- 93\\ \midrule
Summerscales (2014) & 76 & .. \\
BANNER (baseline)\\ \cite{summerscales14thesis} & 64 & .. \\
de Brujin (2008) & 80 & .. \\
Hansen (2008) & 97 & .. \\
Xu (2007) & 92.5 & .. \\
\bottomrule

\end{tabular} 
\caption{\label{tab1} System accuracies for different feature combinations, and the 95\% confidence intervals. - represents leave-out feature scores. Lower part of the table presents the performances of several other systems.}
\end{table}

\subsection{Error analysis}
Of the 50 documents in the test set, our approach made only 6 mistakes, despite the small amount of training data utilised. We analysed these 6 cases, and discovered that three of the mistakes were genuine false positives (\emph{i.e.}, the wrong candidate was chosen). For example, consider the following sentence:
\begin{quote}
 \emph{`Between 1996 and 2001, 1477 patients from 70 hospitals in 14 countries ...'}
 \end{quote} 
Because of the close cluttering of numbers, a number of candidates have almost identical features, which leads to mis-classification (our classifier selected `70' as the size). The candidate generation process, however, is able to identify the right candidate. The three remaining mis-classifications are caused by the fact that the sizes are not mentioned as single numbers in the abstracts. Consider the following sentence, for example: 
\begin{quote}
\emph{Patients ... were randomized to 3 months' treatment with either INH before meals (n = 76) or rosiglitazone 4 mg twice a day (n = 69) ...} 
\end{quote}
In this sentence, the total number of trial participants is not mentioned as a single token. As such, the candidate generation process is not able to generate the right candidate for the machine learning classifier in the next step. Considering the fact that 50\% of the errors are caused by this factor, it will be useful to look into this problem, and perhaps modify the candidate selection process so that such mentions can be detected and passed on to the machine learning algorithm for further processing. 

\section{Conclusions}
In this paper, we have presented a system for extracting sample size information from randomised controlled trials. The technique relies on a hybrid framework--- rules are first applied to extract potential size candidates, and a supervised learning algorithm is then applied to select a single candidate from each document. Our approach shows good performance using a very small training set. Importantly, our feature-rich approach can be applied to a variety of other information extraction tasks, particularly those that are related to size information (\emph{e.g.}, detecting population groups, detecting interventions and comparisons), as such important concepts tend to occur close size mentions. The intent of this approach is to act as an extensible module that can be employed in complex, automated NLP tasks such as question answering, summarisation, and evidence appraisal. 

In the future we will attempt to use these techniques to develop a platform for extracting other crucial information from trial abstracts, such as population groups (\emph{e.g.}, men, women, adolescents, diabetic elderly women, and so on), trial completion size, interventions, doses, and durations. We will also explore the use of larger annotated sets. Finally, from a more technical perspective, we will explore the use of distributional word representations to improve the performance of our system. 

\section*{Acknowledgements}
This study was partially funded by Macquarie University and CSIRO Australia during the author's PhD candidacy. 

\bibliographystyle{acl}
\bibliography{alta15.bib}
\end{document}